\documentclass[letterpaper, 10 pt, conference]{Include/ieeeconf}  % Comment this line out if you need a4paper

\IEEEoverridecommandlockouts                              % This command is only needed if 
\usepackage{mathptmx}       % selects Times Roman as basic font
\usepackage{courier}        % selects Courier as typewriter font
\usepackage{type1cm}        % activate if the above 3 fonts are
\usepackage{amsmath}
\usepackage{amssymb}
\usepackage{makeidx}         % allows index generation
\usepackage{graphicx}        % standard LaTeX graphics tool
\usepackage{multicol}        % used for the two-column index
\usepackage[bottom]{footmisc}% places footnotes at page bottom
\usepackage{enumerate}
\usepackage[all]{xy}
\usepackage{cite}
\usepackage[ruled]{algorithm}
\usepackage[noend]{algpseudocode}
\usepackage[usenames, dvipsnames]{color}
\usepackage{bm}
\usepackage{mathrsfs}
\usepackage{graphicx}
\usepackage{epsfig}
\usepackage{pdfpages}
\usepackage{wrapfig}
\usepackage{lipsum}
\usepackage{times}
\usepackage{setspace}
\usepackage{cancel}
\usepackage{wrapfig}
\usepackage[rightcaption]{sidecap}
\usepackage{adjustbox}
\usepackage{xspace}
\usepackage{float}
\usepackage[pagebackref=true,breaklinks=true,letterpaper=true,colorlinks,bookmarks=false]{hyperref}
\usepackage{subfigure}

\newcommand{\ignore}[1]{}

\newcommand{\norm}[1]{\left\Vert#1\right\Vert} % Norm
\newcommand{\abs}[1]{\left\vert#1\right\vert} % Absolute value
\newcommand{\bbm}{\begin{bmatrix}}
\newcommand{\ebm}{\end{bmatrix}}
\newcommand{\bma}[1]{\left[\begin{array}{#1}}
\newcommand{\ema}{\end{array}\right]}

\DeclareMathAlphabet{\mbf}{OT1}{ptm}{b}{n}
\newcommand{\mbs}[1]{{\boldsymbol{#1}}}
\newcommand{\beq}{\begin{equation}}
\newcommand{\eeq}{\end{equation}}
\newcommand{\bdis}{\begin{displaymath}}
\newcommand{\edis}{\end{displaymath}}
\newcommand{\beqarray}{\begin{eqnarray}}
\newcommand{\eeqarray}{\end{eqnarray}}
\newcommand{\beqarraynn}{\begin{eqnarray*}}
\newcommand{\eeqarraynn}{\end{eqnarray*}}

\newcommand{\p}{\partial}

\newcommand{\trans}{{\ensuremath{\mathsf{T}}}} % transpose
\newboolean{draft-mode}
\setboolean{draft-mode}{true}
\newcommand{\sidenote}[1]{\ifthenelse{\boolean{draft-mode}}{\marginpar{\tiny\raggedright\textsf{\hspace{0pt}#1}}}{}}
\DeclareRobustCommand{\arnote}[1]{\ifthenelse{\boolean{draft-mode}}{\textcolor{blue}{\textbf{AR: #1}}}{}}
\DeclareRobustCommand{\ernote}[1]{\ifthenelse{\boolean{draft-mode}}{\textcolor{cyan}{\textbf{ER: #1}}}{}}
\DeclareRobustCommand{\fhnote}[1]{\ifthenelse{\boolean{draft-mode}}{\textcolor{red}{\textbf{FH: #1}}}{}}
\DeclareRobustCommand{\sdnote}[1]{\ifthenelse{\boolean{draft-mode}}{\textcolor{green}{\textbf{SD: #1}}}{}}

\newcommand{\myparagraph}[1]{\vspace{0.05in}\noindent\textbf{#1}}

\title{\LARGE \bf
Tactile Dexterity: Manipulation Primitives with Tactile Feedback}

\author{
    \authorblockN{Francois R. Hogan, Jose Ballester, Siyuan Dong, and Alberto Rodriguez}
    \authorblockA{
    Massachusetts Institute of Technology --- Department of Mechanical Engineering\\
    \href{https://mcube.mit.edu/research/tactile_dexterity.html}{https://mcube.mit.edu/research/tactile\underbar{ }dexterity.html}}}

\begin{document}
\twocolumn[{%
\renewcommand\twocolumn[1][]{#1}%
\maketitle
\begin{center}
    \centering
    \vspace{-0.35in}
    \includegraphics[width=17cm]{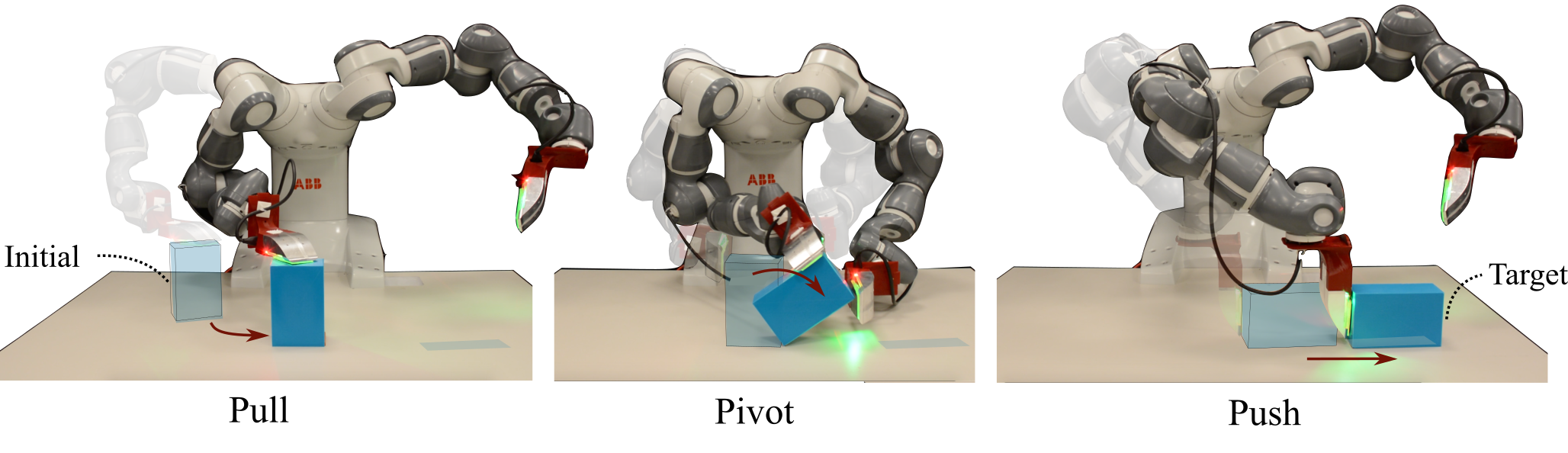}
    \vspace{-0.15in}
    % \captionof{figure}{A dual arm robot manipulates an object from an initial pose to a target configuration. 
    % The manipulation task combines 3 primitive actions to achieve the task: pull the object to the middle of the table, pivot the object, and push it to its target location.}
    \label{fig:lever_experiment_OL}
\end{center}%
}]

\thispagestyle{empty}
\pagestyle{empty}

\begin{abstract}
This paper develops closed-loop tactile controllers for dexterous robotic manipulation with a dual-palm robotic system. \textit{Tactile dexterity} is an approach to dexterous manipulation that plans for robot/object interactions that render interpretable tactile information for control. We divide the role of tactile control into two goals: 1) control the contact state between the end-effector and the object (contact/no-contact, stick/slip) by regulating the stability of planned contact configurations and monitoring undesired slip events; and 2) control the object state by tactile-based tracking and iterative replanning of the object and robot trajectories.

Key to this formulation is the decomposition of manipulation plans into sequences of manipulation primitives with simple mechanics and efficient planners. 
We consider the scenario of manipulating an object from an initial pose to a target pose on a flat surface while correcting for external perturbations and uncertainty in the initial pose of the object. We experimentally validate the approach with an ABB YuMi dual-arm robot and demonstrate the ability of the tactile controller to react to external perturbations.
\end{abstract}
% \vspace{5mm}

% \noindent Keywords: Dexterous Manipulation, Tactile Sensing, Reactive Control

\section{Introduction}
\label{sec:introduction}

\let\thefootnote\relax\footnotetext{This work was supported by the Amazon Research Awards (ARA) and the Toyota Research Institute (TRI).}
%This article solely reflects the opinions and conclusions of its authors and not of Amazon and TRI.
\let\thefootnote\relax \footnotetext{Corresponding author: Francois Hogan \texttt{$<$fhogan@mit.edu$>$}
}

\let\thefootnote\relax\footnotetext{Project website: \href{https://mcube.mit.edu/research/tactile_dexterity.html}{https://mcube.mit.edu/research/tactile\underbar{ }dexterity.html}}

\let\thefootnote\relax\footnotetext{Project video: \href{https://youtu.be/f59FoS-hV7c}{https://youtu.be/f59FoS-hV7c}}

This paper studies the use of tactile sensing for dexterous robotic manipulation, or \emph{tactile dexterity}.
Despite the evidence that humans heavily depend on the sense of touch to manipulate objects~\cite{johansson1984roles}, robots still rely mostly on visual feedback. The vision-based approach has been effective for tasks such as pick-and-place \cite{correll2016analysis} but it presents fundamental limitations to accomplish dexterous manipulation tasks that depend on more accurate and controlled contact interactions, such as object reorientation, object insertion, or almost any kind of object use.
%
%,  which  involve more complex physical  interactions. %These executions require planning and reacting to contact geometry, contact force, and contact velocity.
%
%Manipulating an object with dexterity requires  sensing and reacting to contact interactions.  How is the object moving relative to the fingers?  Where is the object within the hands?
%

The mechanics of object manipulation are driven primarily by the relative motions and forces at the frictional interfaces between object, end-effectors and environment.
As a result, we believe that tactile sensors, with their ability to localize contact geometry, detect contact motion, and infer contact forces, should be at the center of our manipulation plans.

% Therefore, it seems rational that localizing contact geometry, detecting contact motion, and inferring contact forces, should be at the center of our manipulation plans. Tactile feedback is well suited to enable robotic dexterity.

%When using an object, the relative motion at contact, and the of the contact how is the object moving relative to the fingers?  Where is the object within the hands? The sense of touch drives the manipulation process by localizing the contact geometry, inferring interaction forces, and detecting local motions of the object.  
%
%Whereas vision allow us to generally perceive the properties of an object, it is the sense of touch that drives the manipulation process. %When does the hand contact the object?

Our approach to \emph{tactile dexterity}, described in Sec.~\ref{sec:approach}, is based on planning for robot/object interactions that render interpretable tactile information for control.
%
%In this paper, we introduce \textit{tactile dexterity}: an approach to dexterous manipulation that plans for interactions rendering interpretable tactile information for control. 
%
%The exploitation of tactile sensing for robotic manipulation has remained a longwidstanding challenge within the robotics community. This can be attributed to difficulties in interpreting   tactile signals and designing reactive policies. 
%
At its core, this is an approach to robotic manipulation that puts tactile feedback at the center to bypass some of its common caveats.
Tactile information is by nature local and, in general, not sufficient to fully describe the state of a manipulated system~\cite{koval2015pose}. Furthermore, the design of feedback policies for systems undergoing physical contact is challenging, even when full state feedback is available~\cite{woodruff2017planning,hogan2018reactive}.

%Understanding how a robotic system should react to physical interactions with the environment requires complex controller designs.  Even under the assumption of full state feedback, this problem has been shown to be challenging due to the nature of contact interactions, which are discontinuous and underactuated \textcolor{blue}{[cite]}. 
%  \begin{figure}[t]
% \centering
% {
% 			\includegraphics[height=6cm]{figures/yumi_levering_compress} 
% 		\label{fig:lever_experiments}
% }
%
% \caption{Dual-arm manipulation with robotic palms using an ABB YuMi robot with robotic palms. Each robotic palm is equipped with a GelSlim tactile sensor.} 
% \label{fig:YuMi_gelsight}
% \end{figure}
%%%%%%%%%%%%%%%%%%
\begin{figure*}[t]
\centering{\includegraphics[width=17.4cm]{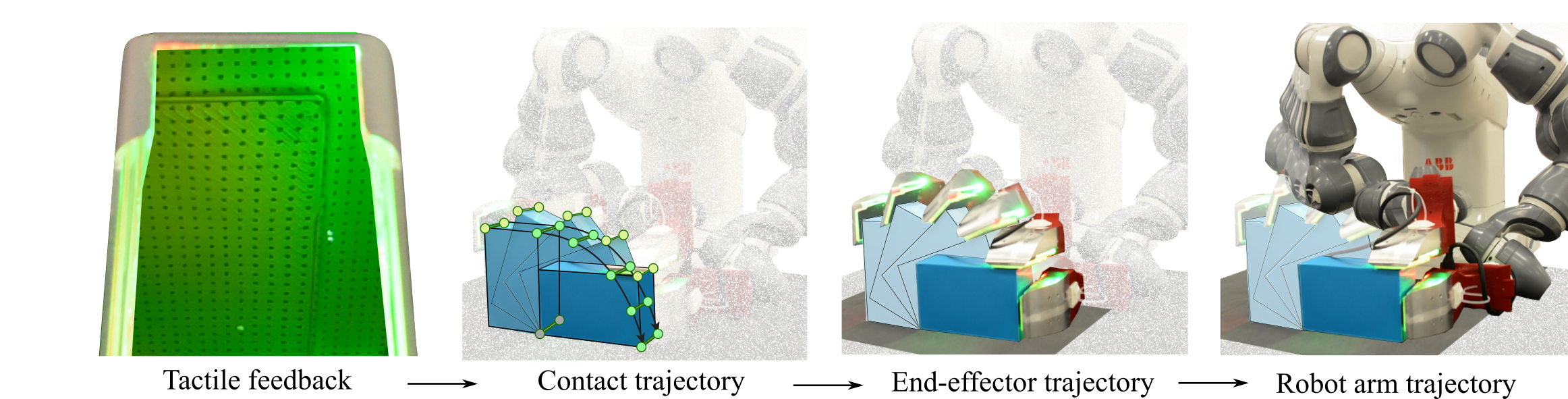}}
\caption{{\small \textbf{Planning to pivot a box.} Pivoting a box involves a hierarchy of planing problems. Tactile sensors allow us to observe and control contact points. The trajectory of contact points is determined by the end-effector pose trajectory, which in turn is generated by the joint trajectory of the robot arm.
% 
% The robot arms move the end-effectors (palms), which in turn generate the trajectory of the contact points that ultimately move the box. Tactile sensors allow us to observe and control those contact points. 
%Our long-term goal is to address this hierarchical planning-control problem. 
We propose an approach to dexterous manipulation that structures the planning problem in a way that when unexpected behavior of the contact points is observed with the tactile sensors, the robot can replan the motion of the contact points, palms, and arms in real-time.}} 
\vspace{-0.15in}
\label{fig:tactile_dexterity}
\end{figure*}
%%%%%%%%%%%%%%%%%%
%

Given these challenges, the key question becomes: How do we structure manipulation planning so that tactile feedback is relevant and convenient?
Our approach to tactile dexterity:
\begin{itemize}
    %\item \textbf{Restricting contact interactions}
    \item \textbf{Targets contact interactions} that produce tactile imprints rich in geometric features (for estimation); and that define dynamic systems with simple mechanics and efficient closed-loop policies (for control). In Sec.~\ref{sec:mechanics} we describe four such manipulation primitives: \texttt{grasp}, \texttt{push}, \texttt{pivot}, and \texttt{pull}.
%
     %\item \textbf{Divide the role of tactile control}
     \item \textbf{Partitions the role of tactile control} into controlling the \emph{contact state} between end-effector and object; and controlling the \emph{object state} in its environment. We describe a formulation for both in Sec.~\ref{sec:tactile_control}.
\end{itemize}

%To address the aforementioned challenges, we structure the manipulation problem to exploit tactile sensing as much as possible. This includes targeting regions of the object that have rich geometric features (observability) and designing manipulation primitives for which we can develop the mechanics and design effective closed-loop manipulation policies (control).

% as a sequencing of manipulation primitives (e.g.,  grasp, push, pivot, pull).  Including this structure as prior knowledge permits us to exploit simple mechanics and efficient planning to develop closed-loop manipulation policies.  

We present a first study of this approach for the scenario of a dual-arm robot equipped with high-resolution tactile-sensing palms (based on GelSlim~\cite{donlon2018gelslim}) as end-effectors, and tasked with manipulating an object on a table-top from an arbitrary initial pose to an arbitrary target pose. An offline graph-search task planner, described in Sec.~\ref{sec:planning}, sequences manipulation primitives, which are   executed in a closed-loop fashion by the robot.
The focus of the experiments described in Sec.~\ref{sec:results} is to evaluate the robustness of the system to external perturbations and to uncertainty in the initial pose of the object.

\section{Related Work}
\label{sec:related_work}

%Tactile sensors have been proposed to enable reactive manipulation capabilities, by evaluating grasp stability, monitoring object slippage, and performing closed-loop control.
This section outlines previous relevant research in exploiting tactile sensing for manipulation control. 

%1. robotic palms
\myparagraph{Slip Control} Veiga~\cite{veiga2015stabilizing} learns a predictor of slippage of an object in a grasp. This model is  used in~\cite{veiga2018hand} to maintain stable multi-fingered grasps under external perturbations, where each finger acts independently to enforce sticking contact. Dong~\cite{dong2019} develops an incipient slip detection algorithm with a GelSlim vision-based tactile sensor~\cite{donlon2018gelslim}, and uses it to design a closed-loop tactile controller that maintains stable grasps in prehensile experiments such as a cap into a bottle with a parallel-jaw gripper. Li~\cite{li2014learning} develops an impedance controller that adapts the grasp on an object  when it is predicted to be unstable. These studies focus on controlling stick/slip interactions but do not explicitly control the trajectory of the manipulated object.  
 %Furthermore, a limitations of the aforementioned is that the control policy is crafted for a particular hand/object configuration and does not easily generalized to various gripper designs. 
%  This paper combines slip detection algorithms with contact physic models to design contact stabilization policies having the ability to generalize to various gripper/object configurations.

\myparagraph{Pose Control} 
Tian~\cite{tian2019manipulation} trains a deep convolutional neural network  to predict the motion of a ball rolled on the ground using a tactile finger directly in tactile space. This model is controlled using sample-based MPC.  A drawback of this approach is its need for large quantities of real-world data, which would be challenging to collect for the rich palm/objects interactions considered in this paper.

Li~\cite{li2014localization} shows that localized object features can be exploited with tactile sensing to recover an accurate estimate of  its pose.  This strategy has been shown to be effective at performing challenging manipulation tasks such as part insertion with small tolerances.  Izatt~\cite{izatt2017tracking} fuses tactile and visual perception by interpreting tactile imprints  as local $3$D pointclouds  within a Kalman filter framework for object pose estimation.   More recently,  Bauza~\cite{bauza2019tactile} develops a tactile based pose estimation algorithm that exploits a high resolution tactile map of the object to localize tactile imprints.
%Localization and Manipulation of Small Parts Using GelSight Tactile
% Sensing

\myparagraph{Dexterous Manipulation}
Erdman  \cite{erdmann1998exploration} first studied the use of robotic palms   as a way to achieve rich manipulation skills  by exploiting the mobility of  robotic manipulators. Researchers have investigated robotic palms for scooping \cite{trinkle1988investigation, trinkle1992stability}, tilting \cite{erdmann1988exploration},  grasping \cite{paljug1993experimental, yun1993object}, and collaborative manipulation \cite{bicchi1995mobility, bicchi1995dexterous}. This paper draws inspiration from  \cite{kaelbling2010hierarchical, Woodruff_ICRA_2017, toussaint2018differentiable} by structuring complex manipulation behavior as a combination of simpler manipulation primitives.

\begin{figure*}[t]%
\centering
%%%%%%%%%%%%
\subfigure[
%Grasp. The robotic palms can align as a parallel jaw gripper to grasp an object. %By securing a stable grasp on the object with both contacts  sticking relative to the object, both arms  move synchronously to achieve the desired object reconfiguration. 
]{%
 \includegraphics[height=3.5cm]{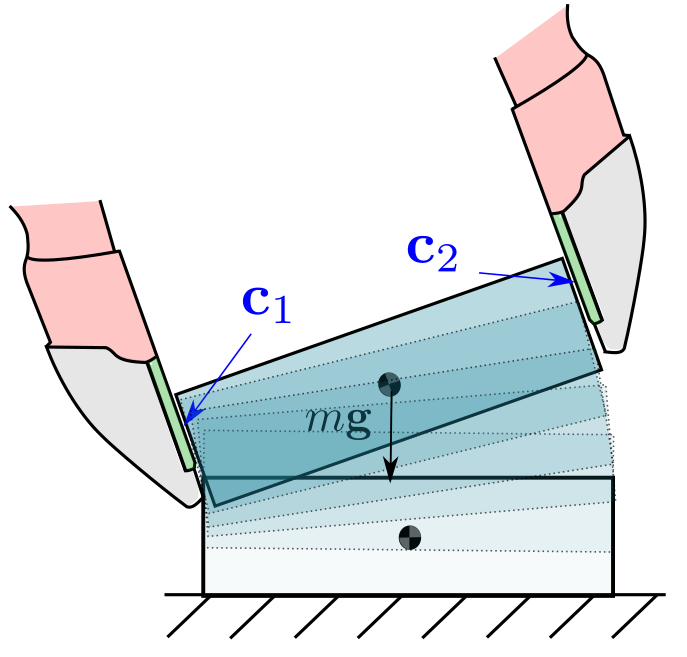}}
 \label{fig:grasping_fbd}%
 \hspace{0mm}
 \subfigure[  %Byorienting itself such that the tactile sensor interacts with anedge of the object, the palm is able to apply both forces andmoments  by  maintaining  line  and  surface  contacts  with  theobject.
 ]{%
 \includegraphics[height=3.5cm]{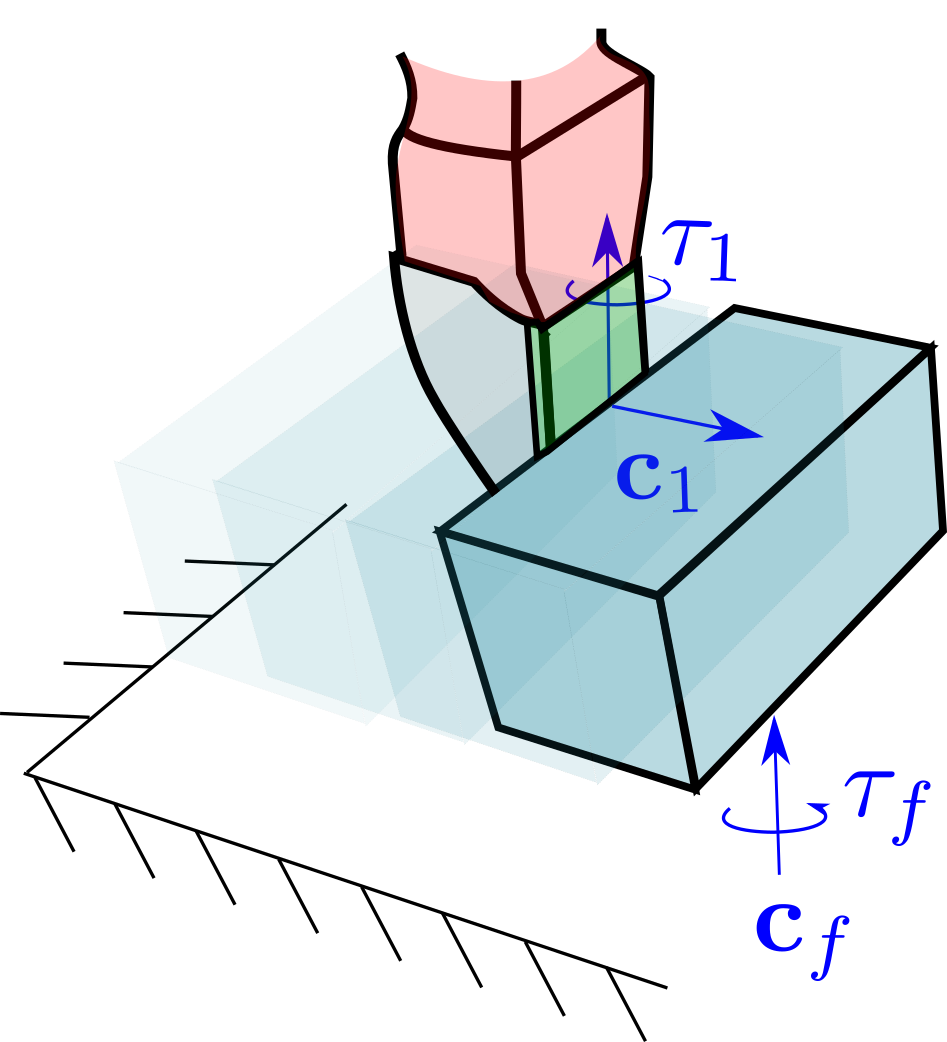}}
 \label{fig:pushing_fbd}%
 \hspace{0mm}
 \subfigure[]{%
 \includegraphics[height=3.5cm]{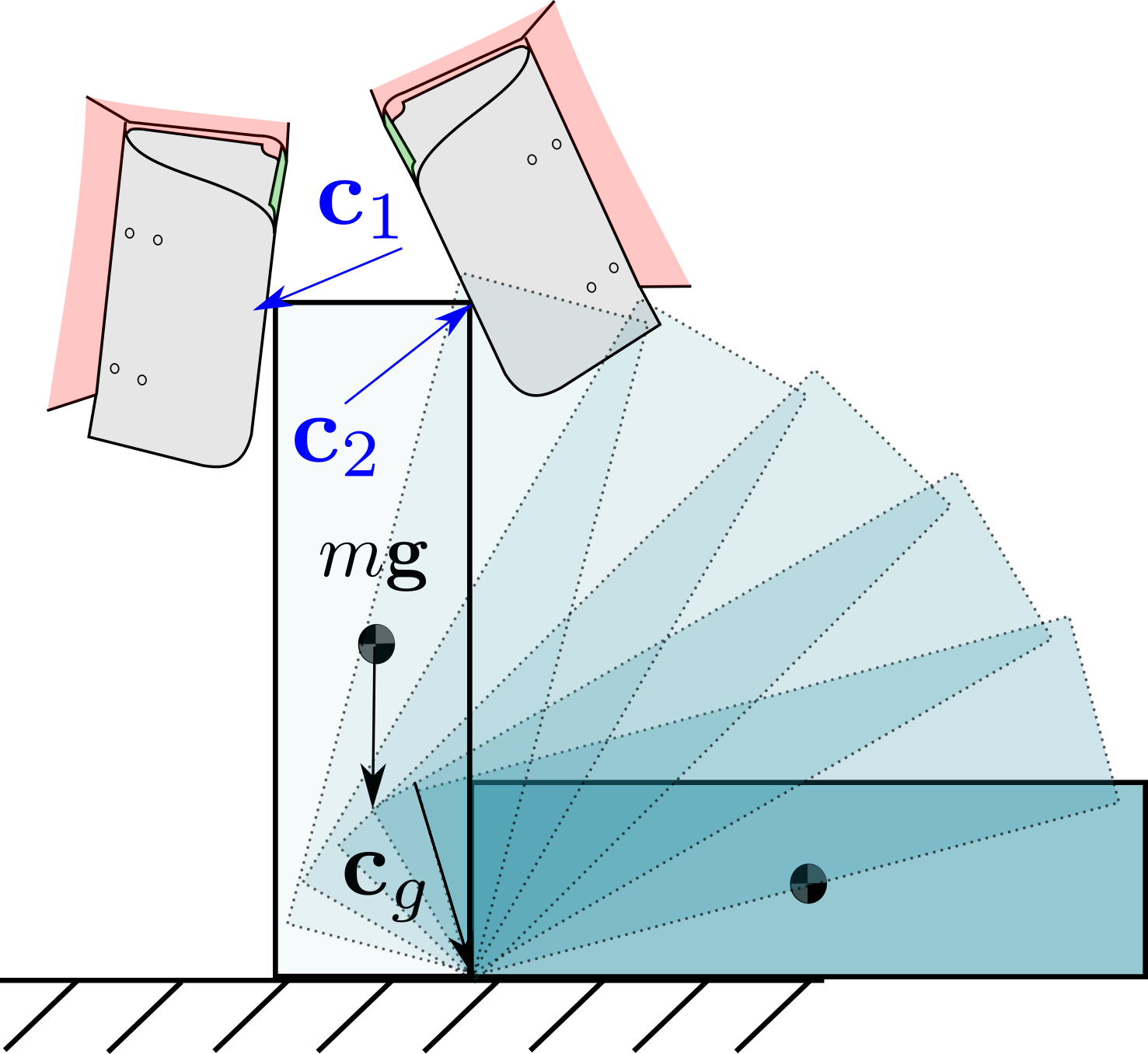}
  \label{fig:levering_fbd}
  }
 \hspace{0mm}
  \subfigure[
  %, provided that the normal applied force is sufficient to prevent translational and rotational slippage.
  ]{%
 \includegraphics[height=3.3cm]{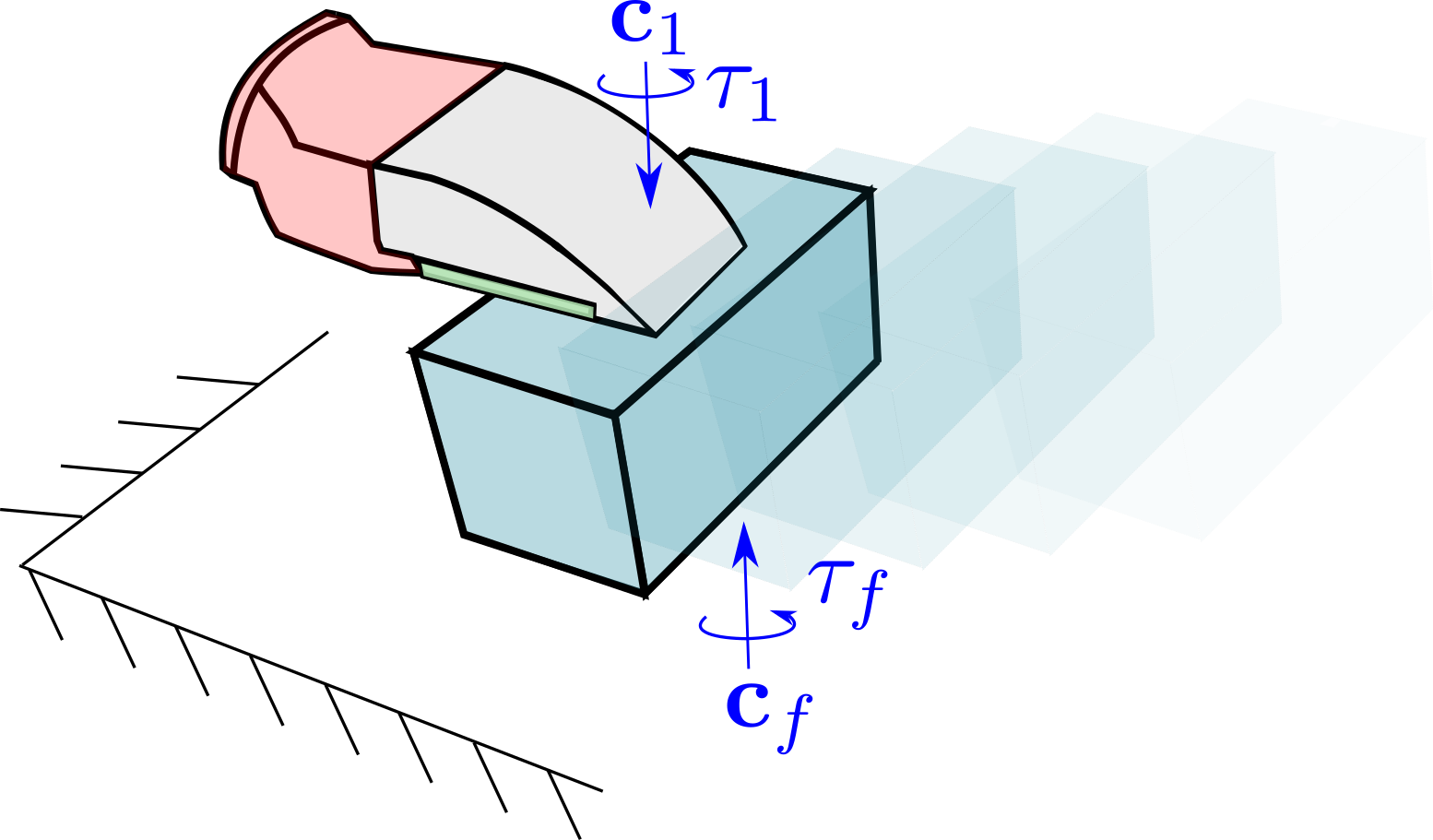}}
 \label{fig:pulling_fbd}%
%%%%%%%%%%%%
\caption{{\small \textbf{Manipulation primitives}. a) \texttt{grasp}: the palms align as a parallel-jaw gripper to grasp an object. b) \texttt{push}: one  palm contacts the object laterally to manipulate it within the plane. c) \texttt{pivot}: the object is rotated about an edge on the table by both palms. d) \texttt{pull}: one palm presses down on the top of the object to slide it within the support plane.}}
\vspace{-0.10in}
\label{fig:manipulation_primitives}
\end{figure*}
%%%%%%%%%%%%

\section{Approach}
\label{sec:approach}

This paper develops closed-loop tactile controllers that enable robust manipulation behavior, where the robot can react to external object perturbations. We propose an approach that divides the role of tactile control into:
%%%%%%%%%%%%%%%%
\begin{itemize}
\item \textbf{Contact state control}. Enforce a desired contact formation (contact/no-contact, stick/slip) between the end effector and the object.
\item \textbf{Object state control}. Control the motion of the object using a contact based state estimator for tactile tracking and iterative replanning of the trajectories of the object, end-effector and robot.
\end{itemize}
%%%%%%%%%%%%%%%%%%
%
This paper considers the scenario of manipulating an object from an initial pose to a target pose on a table top. We consider a robotic platform that is 1) dexterous, where both robotic palms are controlled independently, and 2) tactile sensorized, where each palm is equipped with high-resolution tactile sensors (based on GelSlim technology \cite{donlon2018gelslim}).
We formulate the manipulation problem as a sequencing of manipulation primitives, where each primitive is designed to have a prescribed contact interaction between the robotic palms, the object, and the environment. Fig.~\ref{fig:manipulation_primitives} shows the four manipulation primitive considered in this paper: \texttt{grasp}, \texttt{push}, \texttt{pivot}, and \texttt{pull}.
This structuring of the manipulation problem into simpler behaviors gives us the freedom to design interactions for which we understand the mechanics,  are able to interpret the tactile information, and can develop effective planning algorithms.

\section{Mechanics of Manipulation Primitives}
\label{sec:mechanics}

This section describes the mechanics of the four manipulation primitives in Fig.~\ref{fig:manipulation_primitives}. To model the palm-object-environment interaction we assume:
\begin{itemize}
    \item Known geometry of object, robot, and environment.
    \item Known coefficients of friction. 
    \item Rigid-body interaction. 
    \item Coulomb's friction law.
    \item Quasi-static interaction. 
    \item Surface contacts with uniform pressure distribution.
\end{itemize}
These simplifications will help in designing fast trajectory planning algorithms in Sec.~\ref{sec:planning}. %Our aim is to describe the interactions between the robot, the object, and the environment with simple models useful for control. 
The unmodeled aspects of the interactions (non-uniform pressure distributions, inertial forces, deformation of contact, etc.) are addressed by designing closed-loop tactile controllers in Section~\ref{sec:tactile_control} that react to undesired contact events such as slip.

Assuming quasi-static interactions, force equilibrium dictates that contact forces on the object (applied by the end-effector or environment) are balanced by:
%%%%%%%%
\beq
\sum_{i=1}^C \mbf{G}_i(\mbf{q})^\trans\mbf{w}_i = \mbf{w}_{ext},
\label{eq:static_equilibrium}
\eeq 
%%%%%%%%%%%
where $\mbf{q} = \bma{ccc} \mbf{p}_{o}^{\trans}&
\mbf{p}_{p,l}^\trans&\mbf{p}_{p,r}^\trans\ema^\trans$ is the concatenation of the object pose and the left/right palm poses, $\mbf{w}_i=[\mbf{c}_i^\trans\,\, \mbs{\tau}_i^\trans]^\trans$ is the  applied wrench on the object by the $i^{th}$ contact in the contact frame, $\mbf{w}_{ext}$ is the external force applied by gravity in the world frame,  $\mbf{G}_i$ is a grasp matrix transforming the coordinates of a contact wrench from the contact frame to the world frame \cite{siciliano2016springer}, and $C$ is the number of contacts.

% \begin{algorithm}[b]
%  \caption{Push Planning}
%  \label{alg:push_planning}
% \begin{algorithmic}[1]
% \State \textbf{Given:} Initial object pose $\mbf{p}^o_i$, final object pose $\mbf{p}^o_f$, contact locations $r_i$, robot arm  $A$ \hspace{15mm}
% \Procedure{}{}
% \State $\tau_{o, 1}$, $\tau_{p, 1}$ = MakeContact($\mbf{p}^o_i$, $r_i$, $A$) \label{l:split_joint_start}
% \State $\tau_{o, 2}$, $\tau_{p, 2}$ = DubinsPush($\tau_{o, 1}$, $\tau_{p, 1}$, , $\mbf{p}_f^o$, $A$ )
% \State $\tau_{o, 3}$, $\tau_{p, 3}$ = ReleaseContact($\tau_{o, 2}$, $\tau_{p, 2}$, $A$ )
% \EndProcedure
% \State \Return{Object  trajectory $\tau_o = \text{Concat} (\tau_{o,1}, \hdots, \tau_{o,3})$, \\ \hspace{10mm} Palms  trajectory $\tau_p = \text{Concat} (\tau_{p,1}, \hdots, \tau_{p,3})$}
% \end{algorithmic}
% \end{algorithm}

Contact forces are constrained within the friction cone in accordance to Coulomb's frictional law. Denoting the normal and tangential components of the contact force as $\mbf{c}_i = \bma{cc}f_{n,i}&\mbf{f}_{t,i}^\trans\ema^\trans$, we express Coulomb's frictional law as:
\beqarray
{f}_{n,i}  &\geq& {0}
\label{eq:coulomb_normal}\\
\abs{\mbf{f}_{t,i}}  &\leq& \mu\abs{{f}_{n,i}}.
\label{eq:coulomb_tangential}
\eeqarray
In the case of point contact interactions (\texttt{grasp} and \texttt{pivot}), the contact is unable to sustain frictional moments, implying $\mbs{\tau}_i=\mbf{0}$. For contacts modeled using surface contacts (\texttt{push} and \texttt{pull}), the surface is able to resist a certain amount of frictional moment. We model surface contacts using the limit surface~\cite{goyal1991planar}, which describes the set of forces and moments that can be transmitted through the contact. In practice, we make use of the ellipsoidal approximation to the limit surface introduced in~\cite{howe1996practical} that gives a simple analytical representation of the limit surface. 
A detailed description of the mechanics of similar manipulation primitives can be found in \cite{hogan2018reactive} for non-prehensile pushing, \cite{chavan2018hand} for prehensile pushing, and \cite{hou2018fast} for pivoting. 
 
\section{Tactile Control}
\label{sec:tactile_control}

In this section, we describe tactile-based controllers that give robust behavior to the manipulation primitives.
We use GelSlim
~\cite{donlon2018gelslim}, an optical-based tactile sensor that renders high-resolution images of the  contact surface geometry and strain field, as shown in Fig.~\ref{fig:state_observer}, and in the video \href{https://youtu.be/f59FoS-hV7c}{https://youtu.be/f59FoS-hV7c}.  
 
\begin{figure}[t]%
\centering
{%
 \includegraphics[height=7.5cm]{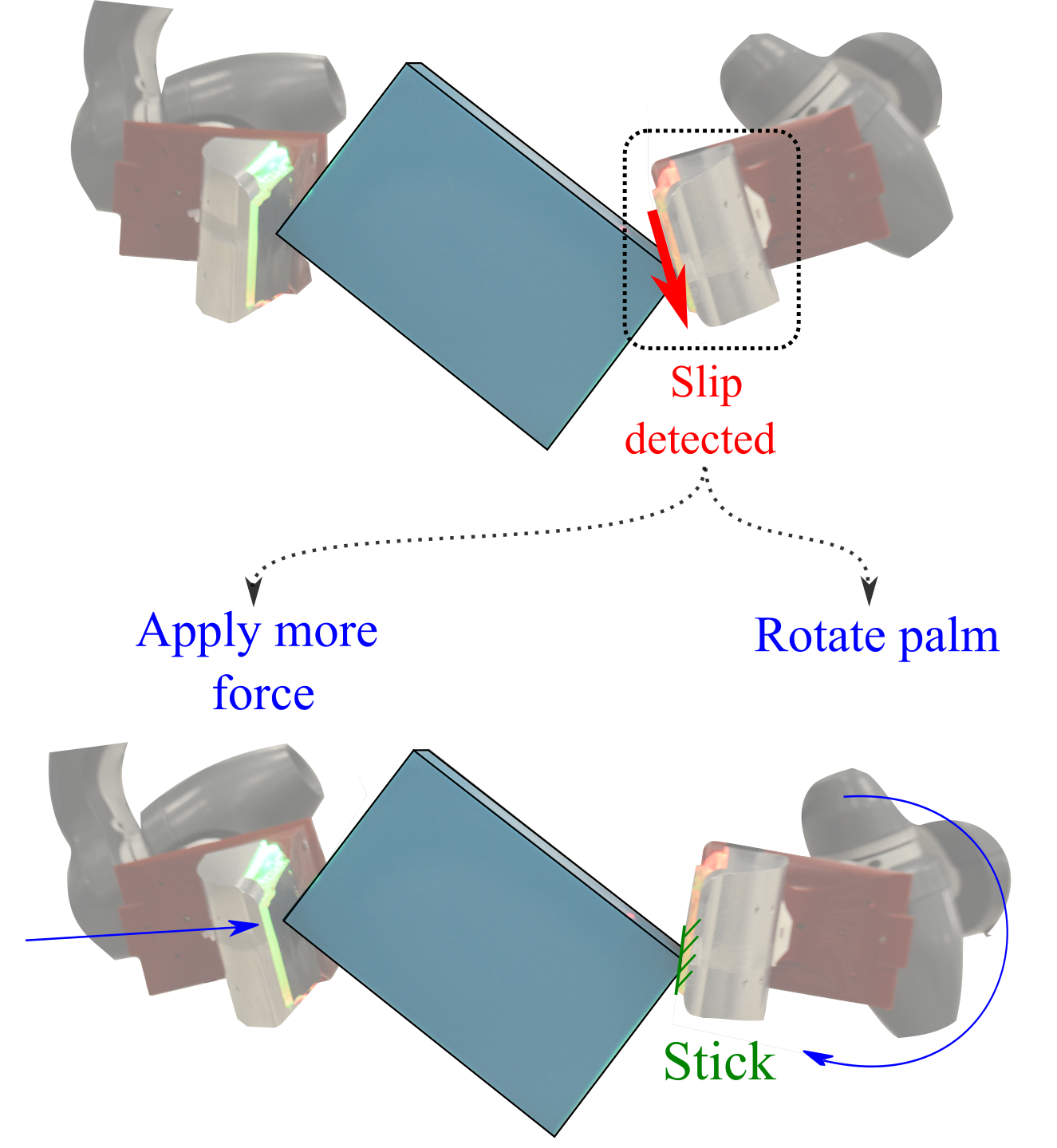}}%
\caption{{\small \textbf{Contact state control}. How should the robot react to undesirable slippage? We design a model based tactile controller that determine locally optimal robot adjustments to recover from contact state deviations.}}
 \label{fig:tactile_controller}
%%%%%%%%%%%%
\end{figure}

A primary goal of the tactile controller is to enforce a desired contact state between the palms and the object. For example, during the pivot maneuver in Fig.~\ref{fig:levering_fbd}, we want both palms to maintain sticking contacts with the corners of the object. By monitoring incipient slippage, we design a controller that engages when necessary to regulate the applied forces on the object to prevent further slip. We refer to this as \textit{Contact State Control (Sec.~\ref{sec:contact_mode_control})}. This is illustrated in Fig.~\ref{fig:tactile_controller}, where the palms rotate and apply additional normal force on the object in reaction to a slip event.
 
An important consequence of undesired slippage is that the position of the object has deviated from the desired trajectory. To address this, we design a tactile-based state estimator, illustrated in Fig.~\ref{fig:state_observer}, that tracks in real-time local features on the object and uses the updated object pose estimate to continuously replan object, palm, and robot arm trajectories for the manipulation primitive. We refer to this as \textit{Object State Control (Sec.~\ref{sec:contact_state_control})}

% We divide the role of tactile control into two objectives: 1) control the contact state between the end effector and the object and 2) control the object state about a nominal trajectory. Designing two independent controllers allows us to fully exploit the available tactile signals, namely slip detection and contact geometry. By monitoring slippage, the contact state controller  regulates the applied forces on the object to enforce a desired contact state between the palms and the object. By monitoring contact geometry, the object state controller estimates the pose of the object and updates the robot's plan to track the desired object trajectory.

% For example, in Fig.~\ref{fig:levering_fbd}, we want both palms to maintain sticking contact with the corners of the object while rotating the object. %By monitoring slippage  during executions, we design a contact mode controller that regulates the applied forces on the object to prevent further slip. This is shown in Fig.~\ref{fig:tactile_controller}, where the contact mode controller rotates the configuration of the palms as well as the applied forces on the object in reaction to a slip event. 
% An important consequence of slippage is that the position of the object has deviated from its nominal trajectory. To address this, we design a tactile-based object state estimator,  shown in Fig.~\ref{fig:state_observer} , that tracks local geometric features on the object and replasn in real time.

\subsection{Contact State Control} 
\label{sec:contact_mode_control}
Each primitive assumes a particular contact formation between the palms and the object. This assumption is likely to be broken as unmodeled perturbations are applied on the system and cause undesired slippage. We design a contact state controller that acts to enforce the planned contact modes by reacting to a binary incipient slip signal $s_i\in\{0,1\}$ at contact $i$, based on  \cite{dong2019}. %In reaction to this signal, we seek local robot adjustments leading the contacts to recover their nominal  mode. %This ensures, for example, that if a modeled coefficient of friction is underestimated, the robot has the ability to adapt its applied force during execution to address this. 
%
%%%%%%%%%%%%
%%%%%%%%%%%%
%
 %We design an optimization based controller that uses the mechanics model developed in Section~\ref{sec:mechanics} to reason about the contact forces. 

\begin{figure}[t]%
\centering
%%%%%%%%%%%%
%[Contact State Control. How should the robot react to undesirable slippage? We design a model based tactile controller that determine locally optimal robot motions to recover from contact state mistakes. In this example, we see that in reaction to detected slip in the right palm, the hand should rotate clockwise to force the reaction force to move towards the center of the friction cone.
%]
 \includegraphics[height=3.3cm]{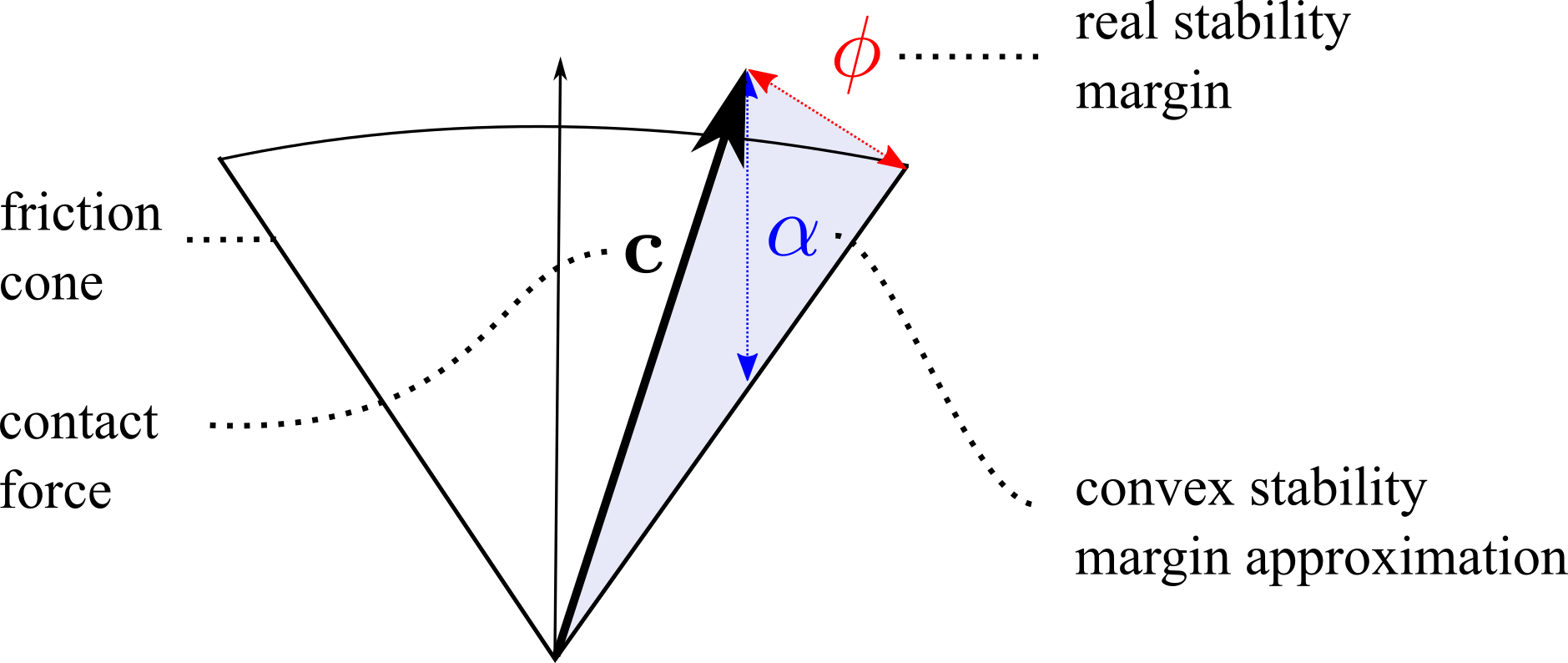}
\caption{{\small The stability margin $\phi$ evaluates how close a contact is to the slipping boundary. The goal of the contact state controller is to maximize/minimize the stability margin to encourage/discourage  slippage. }}
 \label{fig:stability_margin}%
%%%%%%%%%%%%
\end{figure}

Coulomb friction states that slippage occurs when the contact force lies on the boundary of the friction cone, as shown in Fig.~\ref{fig:stability_margin}. Given an undesired slippage signal, we find local palm configuration and force adjustments that result in a more stable sticking solution (i.e. contact forces nearer to the center of the friction cone). The stability margin $\phi$ in Fig.~\ref{fig:stability_margin} is defined as the shortest distance from the contact force to the friction cone boundary and evaluates the risk of slipping of a particular contact. The goal of the controller is to find robot adjustments that either maximize (for sticking) or minimize (for sliding) the stability margin of a contact to enforce a desired contact mode.
In this paper we develop primitives that restrict to sticking contacts between the gripper and the object.

Given the slippage signal ${s}_i\in\{0,1\}$  at each contact and the current robot pose configuration $\mbf{q}_p = [\mbf{p}_{p,l}^\trans\,\,\mbf{p}_{p,r}^\trans]^\trans$, we search for a robot palm pose adjustment $\Delta \mbf{q}_p$ and contact force $\mbf{w}_i$ that maximize the stability margin of a particular contact $i$ (indicated by the weight $\beta_i$) by solving:
%%%%%%%%%%%%%%%
\beq
% \hspace{-5mm}
\begin{aligned}
& \underset{\Delta\mbf{q}_p, \mbf{w}_i}
{\text{max}}
& &  \beta_i \phi_i \\
 \text{s.t.} 
 &&&\sum_{i=1}^C \mbf{G}_i(\mbf{q}_p + \Delta\mbf{q}_p)^\trans\mbf{w}_i = \mbf{w}_{ext} \\
&&&f_{n,i} \geq 0\\
&&& \norm{\mbf{f}_{t,i}}_2 \leq \mu \abs{f_{n,i}}\\
\end{aligned}
% \vspace{-1mm}
\label{eq:nonlinear_tactile_controller}
\eeq
%%%%%%%%%%%%%%%%
with $\mbf{G}_i$, $\mbf{q}_p$, 
%$\mbf{w}_i$, 
$\mbf{w}_{ext}$ as defined in Section~\ref{sec:mechanics} and where $\phi$ is the stability margin shown in Fig.~\ref{fig:stability_margin}. The notation $\mbf{q}_p + \Delta \mbf{q}_p$ is abused here for simplicity, where we ensure that $\mbf{q}_p + \Delta\mbf{q}_p$ belongs to $SE(3)$ and does not violate the kinematics (i.e. maintaining contact and non-penetration) of the  contact formation. %In practice, this will define a different space of possible robot adjustments for each primitive. 
For example, in Fig.~\ref{fig:levering_fbd}, the palms can press harder and/or pivot about the contact edge. The parameter $\beta_i$ is a weight used to biased the effort of the regulating behavior to contact $i$ based on the slip signal $s_i$. %For example, following a slippage event at contact $i$, the controller would increase $f_{n,i}$. 

The optimization program in ~\eqref{eq:nonlinear_tactile_controller} is non-convex because $\phi$ is nonlinear and because the constraint associated with static equilibrium is bilinear. In \cite{aceituno2017simultaneous}, the surrogate stability margin $\alpha$ illustrated in Fig.~\ref{fig:stability_margin} is proposed as a convex approximation to $\phi$. We relax then the optimization program using the surrogate margin $\alpha$ and linearizing the static equilibrium equation \eqref{eq:static_equilibrium} about the current robot configuration $\mbf{q}_p^\star$ and contact forces ${f}_{n,i}^\star$, $\mbf{f}_{t,i}^\star$ for computational efficiency. The linearized contact state controller becomes:
%%%%%%%%%%%%%%%
\beqarray
\begin{aligned}
& \hspace{-2mm}\underset{\Delta\mbf{q}_p, \mbs{\Delta}\mbf{w}_i, \alpha_i}
{\text{max}}
& &  \hspace{-3mm}\beta_i \alpha_i \\
 \text{s.t.} 
 &&&\hspace{-13mm}\sum_{i=1}^C \left(\left.
 \frac{\p \left({G}_i \mbf{w}_i\right)}{\p \mbf{q}_p}\right|_{\star}\mbs{\Delta}\mbf{q}_p
+ \left.
 \frac{\p \left({G}_i \mbf{w}_i\right)}{\p \mbf{w}_i}\right|_{\star}\mbs{\Delta}\mbf{w}_i \right) = \mbf{w}_{ext} \\
&&&\hspace{-13mm}f_{n,i} \geq 0\\
&&&\hspace{-13mm}\alpha_i \geq 0 \\
&&& \hspace{-13mm}\norm{\mbf{f}_{t,i}}_2 \leq \mu {f_{n,i}}\\
&&& \hspace{-13mm}\norm{\mbf{f}_{t,i}}_2 \leq \mu \abs{f_{n,i} - \alpha_i},
\end{aligned}
\hspace{-15mm}\label{eq:linear_tactile_controller}
% \vspace{-1mm}
\eeqarray
where the symbol $(\cdot)^\star$ is used to  evaluate a term at the nominal configuration. Note that the surrogate margin $\alpha$ is not invariant to scale, i.e., we could artificially increase the margin simply by scaling up the magnitude of all the forces. To regulate this behavior, in practice, we add bound constraints to the allowable $\Delta\mbf{q}$ and $\mbs{\Delta}\mbf{w}$. %Two additional constraints associated with alpha in  Eq.~\ref{eq:linear_tactile_controller} are added to the program to define  $\alpha_i$ consistently with  Fig.~\ref{fig:stability_margin}. 
The optimization program in~\eqref{eq:linear_tactile_controller} takes the form of a convex quadratic program under a polyhedral approximation to the friction cone \cite{stewart1996implicit}.

%%%%%%%%%%%%
%%%%%%%%%%%%

\subsection{Object State Control} 
\label{sec:contact_state_control}

% An important consequence of undesired slippage is that the position of the object has deviated from its nominal trajectory. To address this, we design a tactile-based object state estimator, shown in Fig.~\ref{fig:state_observer}, that tracks local geometric features on the object and allows to replan object/palm trajectories in order for the object to reach its target. 

The contact state controller in Sec.~\ref{sec:contact_mode_control} regulates the applied forces on the object to enforce a desired contact mode. This tactile controller reacts to fight against external perturbations but does not have the ability to change the planed trajectory of the object in response to a perturbation. To address this, we design an object state controller running in parallel  tasked with replanning object/palm trajectories to drive the object to  its target location. 

\begin{figure}[t]
\centering
{
			\includegraphics[width=8.3cm]{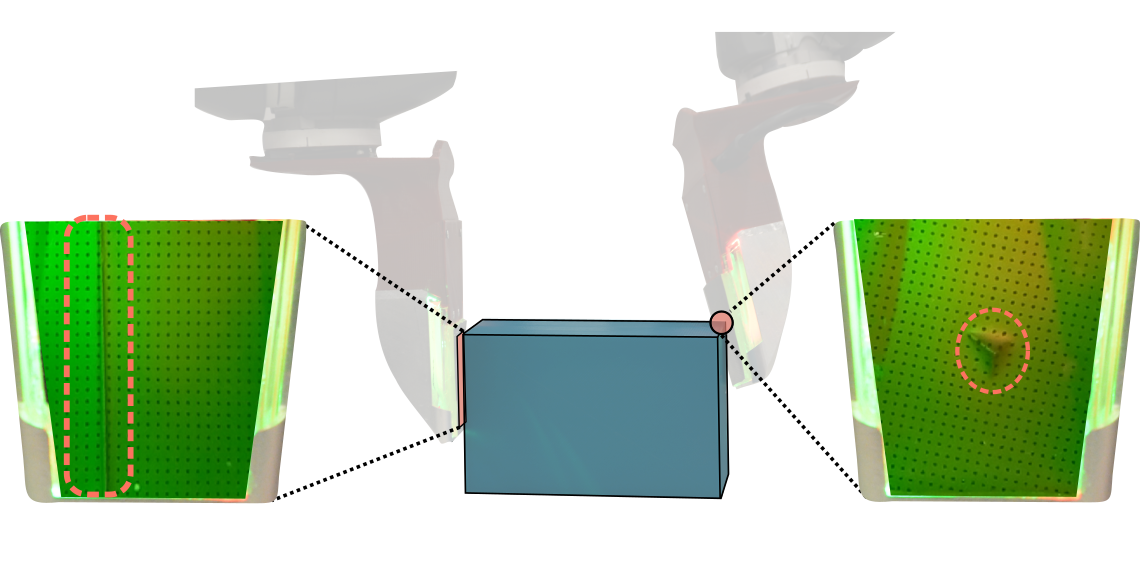} 
}
\caption{{\small \textbf{Tactile object localization}. By localizing descriptive object geometric features (lines, points), we can track the pose of the object.}} 
\label{fig:state_observer}
\end{figure}

In this paper we track two types of features: points (corners of the object) and lines (edges of the object).
% When manipulating an object, we track geometric features in the tactile signal (points, edges, corners) and use them to track the pose of the object. %We use the updated estimate of the object pose to replan the nominal robot/object  trajectories in real-time, allowing the system to adapt to local object pose perturbations.  %
We formulate the tactile object state estimator as an optimization program that updates the pose ${\mbf{p}}_o$ of the object to satisfy the geometric constraints associated with the tactile features, as shown in Fig.~\ref{fig:state_observer}. We quantify the error between the previous and the updated pose estimate using the distance  $d_{TS}(\mbf{p}_o, \mbf{p}_o^\star)$, where $d_{TS}$ is defined as the weighted sum of the Euclidean
metric in $\mathbb{R}^3$ and the great circle  angle metric in ${SO}(3)$ for the respective components~\cite{holladay2016distance}. We enforce that detected lines are collinear with their associated edge on the object mesh and that the sensed points are coincident with the object corner. In addition to the detected geometric constraint, we constrain the estimated object pose to satisfy the geometric constraints consistent with the current manipulation primitive.
For example, for the \texttt{pull} primitive, we enforce that the bottom surface of the object is in contact and aligned with the table top. 

The estimated object pose is used to update the nominal robot palm pose trajectory which allows the robot to adapt to local object perturbations. This is further described in Sec.~\ref{sec:trajectory_planning}.  
%
% Given the previous object pose estimate $\mbf{p}_o^\star$, the detected line ${l}_j$ with associated object edge ${l}^\star_j$ and detected point $x_{k}$ with associated object corner $x_{k}^\star$ solve
% %%%%%%%%%%%%%%%
% \beqarray
% \begin{aligned}
% & \underset{\mbf{p}_o}
% {\text{min}}
% & &  d_{TS}(\mbf{p}_o, \mbf{p}_o^\star) \\
% %
% %
% & \text{s.t.}
% & & \angle (l_j, {l}_j^\star) = 0\\
% &&& {dist} (l_j, {l}_j^\star) = 0\\
% &&& {dist} (x_k, {x}_k^\star) = 0\\
% &&& \mbf{p}_o \in \mathcal{C}_p
% \end{aligned},
% \label{eq:pose_estimation}
% \eeqarray
% %%%%%%%%%%%%%%%%
% where the distance $d_{TS}(\mbf{p}_1, \mbf{p}_2)$ between two poses is defined as the weighted sum of the Euclidean
% metric in $\mathbb{R}^3$ and the great circle  angle metric in ${SO}(3)$ for the respective components \cite{holladay2016distance}.
% The symbol $({\cdot})^\star$ is used to denote the location of the detected feature on the object mesh evaluated at the previous pose $\mbf{q}_o^\star$. For example, the detected line $l_j$ has the associated edge $l_j^\star$ on the object mesh. The symbols $\angle ()$ and $dist ()$ denote the angle and Euclidean distance between two lines and points, respectively. The constraints in Eq.~\eqref{eq:pose_estimation} enforce  that the detected lines/points in the tactile sensor are colinear/coincident with their parent location on the updated object pose, respectively. 

%%%%%%%%%%%%%%%%%%
%%%%%%%%%%%%%%%%%%

%%%%%%%%%%%%
\begin{figure*}[t]%
\centering
%%%%%%%%%%%%
% \subfigure[Pulling] %[Contact State Control. How should the robot react to undesirable slippage? We design a model based tactile controller that determine locally optimal robot motions to recover from contact state mistakes. In this example, we see that in reaction to detected slip in the right palm, the hand should rotate clockwise to force the reaction force to move towards the center of the friction cone.
%]
{%
 \includegraphics[width=18cm]{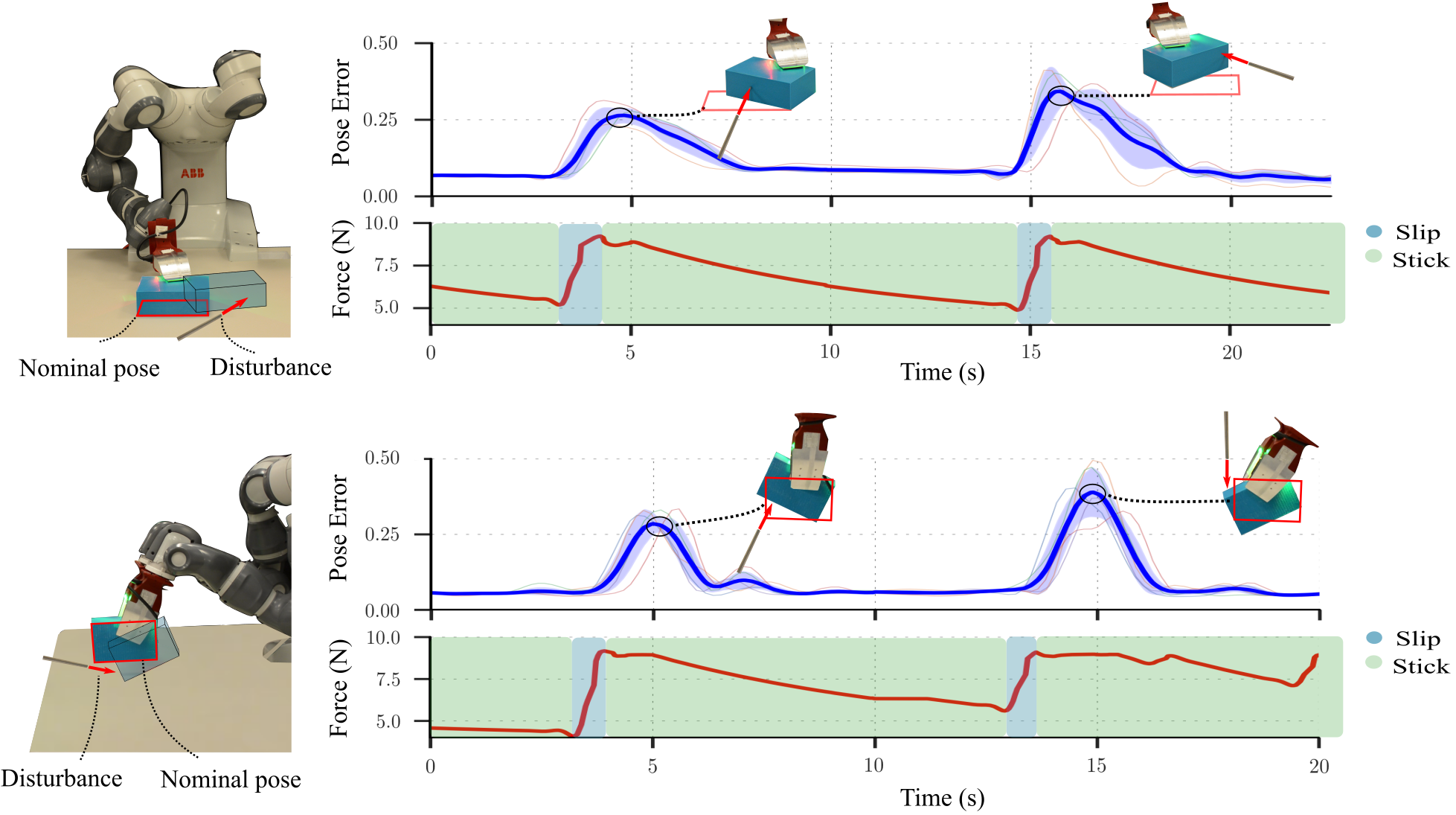}}
 \label{fig:regulation_experiments
  }
% \vspace{-8mm}
% \subfigure[Grasping]{
%   \includegraphics[width=13cm]{figures/grasp_regulation_nominal.png}
%   \label{fig:grasping_regulation}
%  }
%%%%%%%%%%%%
\caption{{\small \textbf{Closed-loop evaluation of the tactile controller}. We consider the task of maintaining the object in a stationary pose under external perturbations. When perturbed, the contact state controller increases the normal force on the object to prevent slippage while the contact state controller replans robot motions to bring the object back to its nominal pose.}}
\label{fig:regulation_experiments}
\end{figure*}
%%%%%%%%%%%%

\section{Planning}
\label{sec:planning}

We formulate the manipulation planning problem as a sequencing of manipulation primitives, as described in Sec.~\ref{sec:mechanics}. This section briefly describes an approach for how to 1) sequence the manipulation primitives to achieve the task and 2) plan robot trajectories within a manipulation primitive a to achieve a desired object transformation.

\subsection{Search for Primitive Sequence}
\label{sec:task_planning}
We formulate the search for the primitive sequence as a graph search problem. We adapt the regrasp graphs developed in~\cite{wan2015improving, hou2018fast} to include a broader set of manipulation primitives. The nodes of the manipulation graph represent possible object stable placements  and the edges represent manipulation primitive actions transforming the object from a stable placement to another. We use Dijkstra's algorithm to search for the shortest path within the constructed graph achieving the desired pose to pose reconfiguration. In the experiment in the title figure, the primitive sequence \texttt{pull$\rightarrow$pivot$\rightarrow$push} is found to achieve the desired object transformations. We refer the reader to~\cite{wan2015improving} and the project website \href{https://mcube.mit.edu/research/tactile_dexterity.html}{https://mcube.mit.edu/research/tactile\underbar{ }dexterity.html} for more details on the graph search planning framework.

\subsection{Trajectory Planning}
\label{sec:trajectory_planning}

After the graph search planner has determined a sequence of primitives, we compute  robot/object trajectories independently for each primitive. We rely on primitive specific planners 
%that return a robot/object pose trajectory satisfying a desired object reconfiguration. 
for which we require that can regenerate trajectories at high frequency to exploit the timely estimated of object pose from the tactile object tracker from  Section~\ref{sec:contact_state_control}. %As such, the function $ManipulateObject$ developped for each primitive is optimized for fast execution and can be run in real-time to consume the $90$ Hz frequency associated with the tactile sensor.
 
%  Given a desired object reconfiguration and specified primitive, Algorithm~\ref{alg:primitive_planning} computes the trajectory of object poses  $\mbs{\tau}_o$ and palm poses $\mbs{\tau}_p$  achieving the object transformation.  %During the execution of a primitive, we require that the planner can recompute trajectories at fast frequencies in order to react to tactile information in a timely fashion. The functions $TransformObject$, $DubinsPush$, and $PivotObject$ are optimized for fast execution and can be run fast enough to consume the $90$ Hz frequency associated with the tactile sensor.
% In addition to the dynamics of each primitive  described by Eq.~\eqref{eq:static_equilibrium}, we exploit the kinematic constraints of a particular primitive to develop efficient planning algorithms. 
%Algorithm~\ref{alg:primitive_planning} describes the structure of a manipulation primitive planner, that returns a palm/object trajectory given the current pose of the object, the target pose, and the contact locations of the palms on the object. 
For \texttt{grasp} and \texttt{pull}, we plan object motions that kinematically stick to the palms. For these, we directly interpolate between the initial and final poses of the object to determine the palm pose trajectory. For \texttt{pivot}, we compute an interpolated object trajectory between the initial and final poses about a specified center of rotation.  Solving~\eqref{eq:static_equilibrium}, we find the palm pose relative to the object that maintains a sticking interaction at all contacts. For \texttt{push}, we use a Dubins' car planner that computes the time-optimal trajectory  connecting  the initial and final object configurations with a single push based on~\cite{zhou2017pushing}. %This pushing planner is largely based on results from \textcolor{blue}{[cite]} that found that planning for pushing motions can be posed as a kinematic problem by restricting object motions to remain within the pusher motion cone. 
We consider pushes on all sides of the object and execute the trajectory with the shortest path.

\section{Results}
\label{sec:results}

We validate our approach to \textit{tactile dexterity} on an ABB YuMi dual-arm robot with real experiments visualized in \href{https://youtu.be/f59FoS-hV7c}{https://youtu.be/f59FoS-hV7c} that evaluate the ability of the tactile controller to handle external perturbations on a table top manipulation task.

The title figure shows snapshots of an experiment where the robot manipulates an object from an initial pose  $\mbf{q}_0 = [0.3,\,\,
-0.2,\,\,
 0.07,\,\,
0.38,\,\,
 0.60,\,\,
 0.60,\,\,
0.38]^\trans$ to a target pose $\mbf{q}_f = [ 0.45,\,\,
0.3,\,\,
0.045,\,\,
0.0,\,\,
0.71,\,\,
0.0,\,\,
0.71 ]^\trans$. To achieve the task, the robot follows the  sequence: \texttt{pull} the object to the middle of the table, \texttt{pivot} the object to its target placement, and \texttt{push} it to its target location. The initial \texttt{pull} primitive is necessary to move the object to a location that allows the robot to perform a pivot manoeuvre with well defined inverse kinematics and that avoids collisions with the environment. 

In Fig.~\ref{fig:regulation_experiments}, we evaluate the closed-loop performance of the tactile controller in Section~\ref{sec:tactile_control} on individual primitives. We illustrate the regulation task of maintaining an object in a stationary pose for the \texttt{pull} and \texttt{grasp} primitives. The regulation task allows to better visualize the reactive capabilities of the controller without loss of generality. In each experiment, we apply two successive impulsive forces on the object and evaluate  the stabilizing capabilities of the tactile controller.
Figure~\ref{fig:regulation_experiments} plots the mean and standard deviation of the error between the object's desired and measured poses for $5$ consecutive experiments. In both cases, the controller quickly reacts to the disturbance by 1) detecting slippage events at the contact interfaces and 2) tracking the pose of the object using the detected object edge in the tactile signal. First, the applied normal force is increased following \eqref{eq:linear_tactile_controller} in reaction to the detected slippage at the contact interface. Second, the robot replans its trajectory from updates on the object state and this quickly returns to its nominal pose.  For evaluation purposes, we track the ground truth pose of the object using an Apriltag visual marker~\cite{olson2011apriltag}.% and the pose distance $d_{TS}$ is a weighted sum of the Euclidean metric in $\mathbb{R}3$ and the great  circle  solid  angle  metric  in $SO(3)$ for  the  respective components  [cite].   

\section{Discussion}
\label{sec:discussion}

This paper introduces \textit{tactile dexterity}, an approach to dexterous manipulation that exploits tactile sensing for reactive control. 
By structuring the manipulation problem as a sequence of manipulation primitives that render interpretable tactile information, we enable tactile object state estimation and tactile object control, yielding robust manipulation behavior. 
This requires targeting primitive interactions to those that have simple rigid body mechanics models and efficient planners.
We show the ability of the tactile controllers to modulate contact forces, track the pose of the object, and handle external perturbations by replanning in real-time.

This research relies on a number of important assumptions about the structure of the problem:
\begin{itemize}
    \item Known environment. The manipulation primitives and planning algorithms are tailored to a flat surface environment. 
    \item Manipulation primitives. We use human intuition to develop a library of manipulation primitives that are amenable for planning and control (for controllability). 
    \item Tactile Pose Estimation. We assume that the pose of the object can be reconstructed from tactile observations (for observability).
\end{itemize}
These assumptions facilitate the development of state estimation, planning, and control frameworks capable of exploiting tactile sensing for dexterous manipulation. They constitute both the limitations, as well as strengths of the approach.

We believe that there are natural extensions to generalize this approach to dexterous manipulation that puts tactile feedback at the center:

\myparagraph{General Manipulation Primitives.} 
Developing manipulation primitives with richer contact interactions (contact switches, stick/slip interactions, etc.) will allow more expressive behavior. Due to increased complexity of the mechanics of these primitives, they will likely benefit even more from real-time tactile observations of contact. 
This line of research will requires controller architectures that can reason across hybrid contact switches, e.g. developed in \cite{hogan2016feedback, doshi2019hybrid}. 

Another interesting possibility is to learn manipulation primitives from experience, and defining a primitive, broadly speaking, as a behavior for which the robot can learn a stable controller. This would potentially reduce the need for human intuition. 
%This approach could potentially discover novel manipulation strategies difficult to  engineer.

\myparagraph{Object tracking.} 
A formal approach to characterizing object observability directly from tactile feedback could assist with identifying discriminative features and grasps of the object. Furthermore, combining touch with vision is a natural extension of this research that would allow the robot to recover from larger object disturbances.

\myparagraph{Task-and-Motion Planning}. Dealing with more complex environments and tasks (e.g., pick up the box and put it on a shelf, or inside a closed drawer), would require to integrate the system with a high level task planner that can sequence manipulation primitives, as is commonly done in task and motion planning, while still being mindful of the mechanics of contact \cite{toussaint2018differentiable, holladay2019force}. Ideally, these planners would be reactive and allow the robot to switch primitives in response to unexpected deviations.

{\small\bibliography{icra_ref}
\bibliographystyle{Include/splncs}

\end{document}